\documentclass[conference]{IEEEtran}
\IEEEoverridecommandlockouts

\usepackage{cite}
\usepackage{amsmath,amssymb,amsfonts}
\usepackage{algorithmic}
\usepackage{graphicx}
\usepackage{textcomp}
\usepackage{xcolor}
\def\BibTeX{{\rm B\kern-.05em{\sc i\kern-.025em b}\kern-.08em
    T\kern-.1667em\lower.7ex\hbox{E}\kern-.125emX}}

\ifCLASSOPTIONcompsoc
\usepackage[caption=false, font=normalsize, labelfont=sf, textfont=sf]{subfig}
\else
\usepackage[caption=false, font=footnotesize]{subfig}
\fi

\usepackage{gensymb}
\usepackage{multirow}

\usepackage{multirow}
\usepackage{makecell}
\usepackage{bm}
\usepackage{enumerate}
\usepackage{arydshln}
\usepackage{multirow}
\usepackage{tabularx}
\usepackage{algorithm}
\usepackage{amsthm}
\usepackage{color}
\usepackage{xfrac}
\usepackage{threeparttable}
\usepackage{siunitx}
\usepackage{caption}
\usepackage{hhline}
\definecolor{best}{RGB}{255, 153, 153}
\definecolor{second}{RGB}{255, 204, 153}
\newcolumntype{C}{>{\centering\arraybackslash}X}

\usepackage[pagebackref,breaklinks,colorlinks]{hyperref}
\usepackage[symbol]{footmisc}

\usepackage[capitalize]{cleveref}
\crefname{section}{Sec.}{Secs.}
\Crefname{section}{Section}{Sections}
\Crefname{table}{Table}{Tables}
\crefname{table}{Tab.}{Tabs.}

\begin{document}

\title{\LARGE \bf \vspace{18pt} Place Recognition under Occlusion and Changing Appearance via Disentangled Representations
}

\author{Yue Chen$^{1}$, Xingyu Chen$^{1,\dagger}$ and Yicen Li$^{2}$ 

\thanks{$^{1}$Xi'an Jiaotong University, China.}%
\thanks{$^{2}$McMaster University, Canada.}%
\thanks{$^{\dagger}$Corresponding author's email:{\tt\small xingyu@stu.xjtu.edu.cn}}
}

\maketitle

\begin{abstract}
Place recognition is a critical and challenging task for mobile robots, aiming to retrieve an image captured at the same place as a query image from a database.
Existing methods tend to fail while robots move autonomously under occlusion (e.g., car, bus, truck) and changing appearance (e.g., illumination changes, seasonal variation).
Because they encode the image into only one code, entangling place features with appearance and occlusion features. 
To overcome this limitation, we propose \textit{PROCA}, an unsupervised approach to decompose the image representation into three codes: a place code used as a descriptor to retrieve images, an appearance code that captures appearance properties, and an occlusion code that encodes occlusion content. 
Extensive experiments show that our model outperforms the state-of-the-art methods.
Our code and data are available at \href{https://github.com/rover-xingyu/PROCA}{https://github.com/rover-xingyu/PROCA}.

\end{abstract}


\section{Introduction}
Many problems in the localization of mobile robot focus on recognizing the place, including detecting the loop closure for Simultaneous Localization and Mapping (SLAM) and finding the initial pose for finer 6-DoF pose regression. Due to the properties of low cost, high resolution, and rich color information of cameras, visual place recognition has received extensive concern. For a query image, the goal is to retrieve an image corresponding to the same place from a database of geo-tagged images and approximate the pose of the retrieved image as the query’s pose.

The query and database images are taken under various occlusion and appearance conditions in many scenarios. For example, changing lighting and seasons lead to appearance variation, and dynamic objects contribute towards occlusion, like cars, buses and trucks around robots. Unfortunately, existing techniques generally encode an image into only one code, entangling place features with appearance and occlusion features. As a result, they fail to retrieve the correct images from a database. Some approaches disentangle content and appearance features, but are still affected by dynamic objects.

   \begin{figure}[thpb]
      \includegraphics[width=1\linewidth]{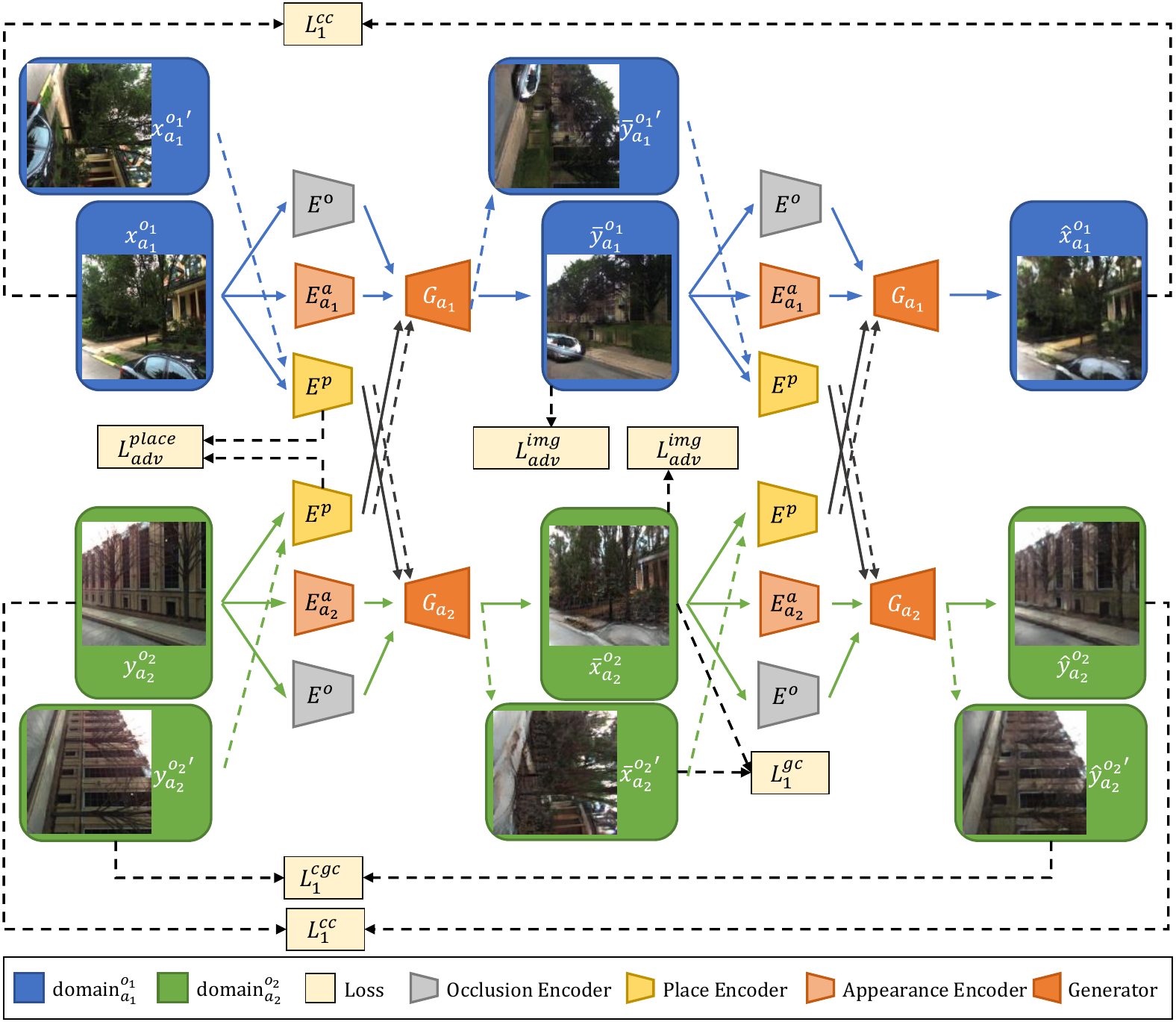}
      \caption{Method overview. We train the PROCA with image adversarial objectives $\mathcal{L}_{\mathrm{adv}}^{\mathrm{img}}$ (appearance adversarial loss $\mathcal{L}_{\mathrm{adv}}^{\mathrm{appearance}}$, occlusion adversarial loss $\mathcal{L}_{\mathrm{adv}}^{\mathrm{occlusion}}$) and latent adversarial loss $\mathcal{L}_{\mathrm{adv}}^{\mathrm{place}}$ to ensure disentangled representations, and cross-cycle consistency loss $\mathcal{L}_{1}^{\mathrm{cc}}$ to learn the mapping between the domain $x_{a_{1}}^{o_{1}}$ and $y_{a_{2}}^{o_{2}}$ with unpaired data. With the proposed geometry consistency loss $\mathcal{L}_{1}^{\mathrm{gc}}$ and the cross-cycle geometry consistency loss $\mathcal{L}_{1}^{\mathrm{cgc}}$, we can further disentangle the place code and the occlusion code by applying a geometric transformation on input images. Benefiting from the disentangled representations, we can use $E^{p}$ to encode a query image to a place code. At test time, the place code can be used as the descriptor to retrieve the image corresponding to the same place from a pre-encoded features database.
      }
      \vspace{-0.5cm}
      \label{fig:framework}
   \end{figure}

\begin{figure*}[thpb]
    \centering
	  \subfloat[CycleGAN\label{fig:CycleGAN}]{
       \includegraphics[width=0.22\linewidth]{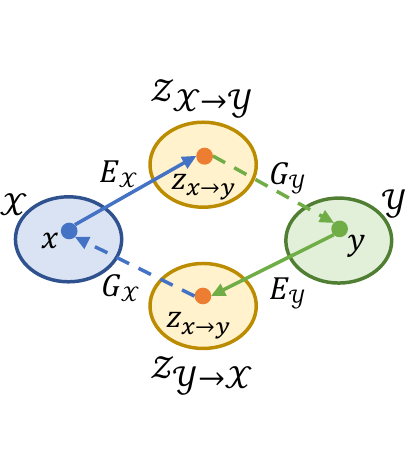}}
    \hfill
	  \subfloat[UNIT, ComboGAN\label{fig:UNIT}]{
        \includegraphics[width=0.22\linewidth]{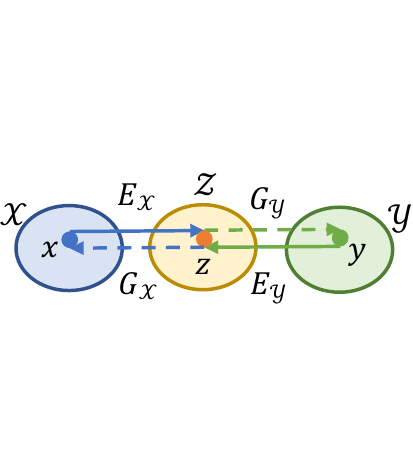}}
    \hfill
	  \subfloat[MUNIT, DRIT\label{fig:MUNIT}]{
        \includegraphics[width=0.22\linewidth]{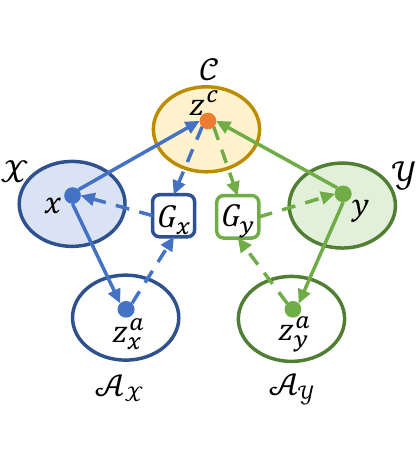}}
    \hfill
	  \subfloat[PROCA\label{fig:PROCA}]{
        \includegraphics[width=0.22\linewidth]{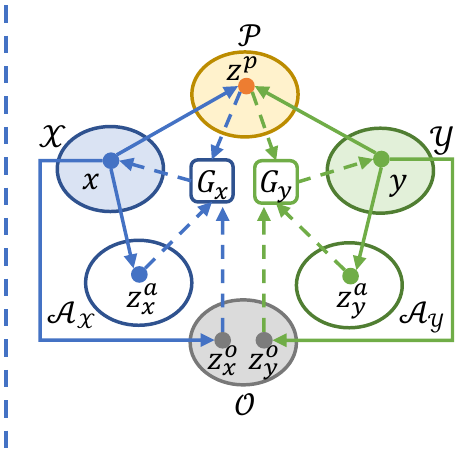}}
	\caption{Comparisons of unsupervised image-to-image translation methods. Denote $x$ and $y$ as images in domain $\mathcal{X}$ (e.g., winter with occlusion) and $\mathcal{Y}$ (e.g., summer without occlusion): (a) CycleGAN maps $x$ and $y$ onto separated latent spaces. (b) UNIT and ComboGAN assume $x$ and $y$ can be mapped onto a shared latent space. (c) MUNIT and DRIT disentangle the latent spaces of $x$ and $y$ into a shared content space $\mathcal{C}$ and an appearance space $\mathcal{A}$ of each domain. (d) We disentangle the latent spaces of $x$ and $y$ into a shared place space $\mathcal{P}$, an appearance space $\mathcal{A}$ of each domain, and a shared occlusion space $\mathcal{O}$.}
	\vspace{-0.5cm}
	  \label{fig:compare} 
	\end{figure*}

In this paper, we propose an unsupervised disentangled representation framework (Figure~\ref{fig:framework}) for Place Recognition under Occlusion and Changing Appearance (PROCA). Our framework makes several assumptions. Firstly we assume that the latent space of an image can be decomposed into a place space, an appearance space and an occlusion space, as shown in Figure~\ref{fig:PROCA}.
We further postulate that images in different domains share a common place space and a common occlusion space but not the appearance space. Because the aligned training image pairs are either difficult to collect (e.g., tuples of images depicting the same place under different conditions) or do not exist (e.g., images 
taken in the identical position under different conditions), we use the unsupervised image-to-image translation task as an auxiliary supervision. 
To translate an image to the target domain, we recombine its place code with an appearance code and an occlusion code from the target domain. During translation, the place code encodes the information that should be preserved, while the appearance code and occlusion code represent the remaining variables that are not contained in the input image.

We disentangle the representations by using a place adversarial loss to encourage the place code not to take domain-specific cues (e.g., edges of buildings), an appearance adversarial loss to facilitate the appearance code to carry appearance properties (e.g., sunlight intensity), an occlusion adversarial loss to stimulate the occlusion code to carry occlusion content (e.g., edges of cars). To tackle the unpaired problem, we first perform a cross-domain mapping to translate the image to another domain by swapping the place codes from unpaired images. We then reconstruct the original image pair by using cross-domain mapping again. Finally, we enforce the consistency between the original and the reconstructed images by leveraging a cross-cycle consistency loss. Moreover, we propose a geometry consistency loss and a cross-cycle geometry consistency loss to further separate the place code and the occlusion code,
which is constructed by applying the cross-domain mapping to the counterpart image pair transformed by our predefined geometric transformation. At test time, we use the disentangled place code to retrieve an image corresponding to the same place from a pre-encoded database.

Extensive experiments demonstrate the effectiveness of our method in handling occlusion and changing appearance and the superiority in localization accuracy compared with state-of-the-art approaches.

\section{RELATED WORKS}

\subsection{Image-to-Image Translation}
Image-to-Image translation aims to learn the mapping from a source image domain to a target image domain. Introduced by Isola et al., Pix2pix~\cite{pix2pix} proposes the first unified framework for image-to-image translation based on conditional generative adversarial networks, using manually labeled image pairs. To train with unpaired data, the CycleGAN~\cite{cyclegan} schemes leverage cycle-consistency to regularize the training, which enforces that if we translate an image to the target domain and back, we should reconstruct the original image, as shown in Figure~\ref{fig:CycleGAN}. UNIT~\cite{UNIT} and ComboGAN~\cite{Combogan} further assume a shared latent space such that corresponding images in two domains are mapped to the same latent code, as illustrated in Figure~\ref{fig:UNIT},

However, a summer scene could have diverse possible appearances during winter due to climate, timing, lighting, etc. For example, a summer landscape of lush greenery may become snow-covered or just leafless in winter. But supposed to enforce the reconstructed image to be the same as the original image by cycle-consistency constraint, the latent code will be entangled with appearance properties, which hampers the place recognition task. Unlike existing works, our method enables image-to-image translation with disentangled representations.

   \begin{figure*}[t]
      \includegraphics[width=1\linewidth]{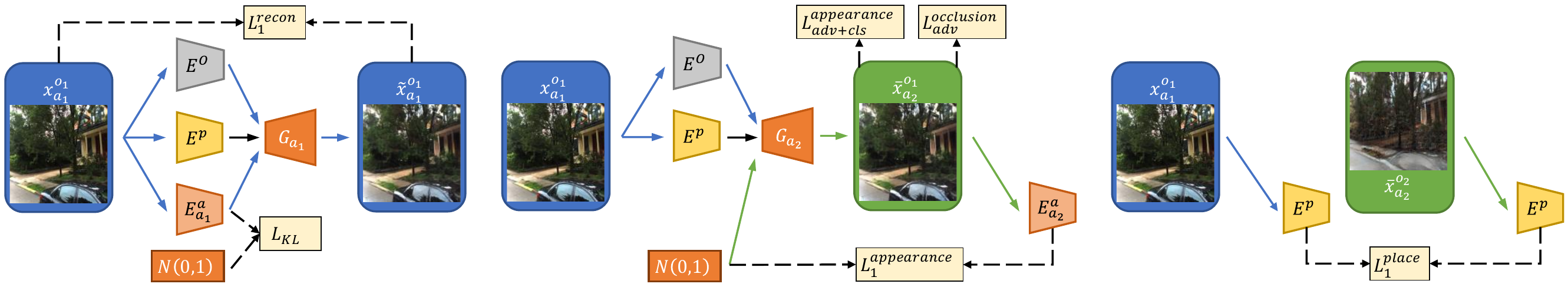}
      \caption{Auxiliary constraints. In addition to the main constraints described in Figure~\ref{fig:framework}, we apply four auxiliary constraints in the training process. The self-reconstruction loss $\mathcal{L}_{\mathrm{1}}^{\mathrm{recon}}$ facilitates training with the auxiliary self-reconstruction task, and the KL loss $\mathcal{L}_{\mathrm{KL}}$ aims to align the appearance representation with a prior Gaussian distribution. The appearance latent regression loss $\mathcal{L}_{\mathrm{1}}^{\mathrm{appearance}}$ enforces the reconstruction of the latent appearance code. The latent place regression loss $\mathcal{L}_{\mathrm{1}}^{\mathrm{place}}$ enforces the reconstruction of the latent place code, which further filters out the occlusion to extract the disentangled place code.
      }
      \vspace{-0.5cm}
      \label{fig:auxiliary_constraint}
   \end{figure*}

\subsection{Disentangled Representations}
Learning disentangled representation refers to modeling the factors of data variations. Previous works leverage labeled data to factorize representations into class-related and class-independent components~\cite{cheung2014discovering,kingma2014semi,mathieu2016disentangling}. Recently, numerous unsupervised methods have been developed ~\cite{infogan,DRIT,MUNIT} to learn disentangled representations. 
As shown in Figure~\ref{fig:MUNIT}, DRIT~\cite{DRIT} and MUNIT~\cite{MUNIT} separate appearance and content components with adversarial and reconstruction objectives. They define "content" as the underlying spatial structure and "appearance" as the rendering of the structure.

However, a place could have many possible contents due to the occlusion caused by dynamic objects such as cars, buses and trucks. It is necessary for us to disentangle place content from occlusion content. 
In our setting, we 
disentangle an image into place, appearance and occlusion representations to facilitate learning place descriptors to retrieve images.

\subsection{Place Recognition and Localization}
Place recognition tackles retrieving the image taken from the same place for a query image, usually used as a key module for the localization pipelines of mobile robots. Typical place recognition includes feature extraction and matching, optionally followed by sequence fusion~\cite{seqslam,siam2017fast} to reduce the false positive rates, among which this paper focuses on the first part.

Conventional methods design features~\cite{lowe2004distinctive,sift,orb}, or statistics of features as global descriptors~\cite{bow,vlad,DenseVLAD,netvlad} to handle a fraction of illumination variation.
However, they have been shown to be heavily affected by domain shifts such as weather, season and occlusion~\cite{Benchmarking}.

To close the domain gap, methods~\cite{porav2018adversarial,todaygan} perform image translation from a source domain(e.g., night) to a target domain (e.g., day) and retrieve the output image through conventional descriptors. 
However, the retrieval effectiveness of these two-stage methods depends on the quality of translated image, which degrades their generalization ability.

By contrast, retrieval with latent features has the advantages of direct, low-cost and efficient learning. Methods~\cite{DIFLFCL,DISAM} are proposed to learn the domain-invariant representation via a ComboGAN-like~\cite{Combogan} architecture. However, based on cycle-consistency constraints, they still try to translate and reconstruct the image from only one content code, which entangles the content code with appearance properties.
Tang et al.~\cite{tang2020adversarial} propose to use adversarial features disentanglement directly on the latent code to separate content and appearance. Nevertheless, the method is yet hampered by occlusion content from dynamic objects, while ours is not.

\section{PROPOSED METHOD}
We propose to learn disentangled representations with the help of image-to-image translation. Given unpaired images $x_{a_{1}}^{o_{1}}$ and $y_{a_{2}}^{o_{2}}$ in two visual domains $\mathcal{D}_{a_{1}}^{o_{1}} \subset \mathbb{R}^{H \times W \times 3}$, e.g., winter $\left(a_{1}\right)$ with occlusion $\left(o_{1}\right)$, and $\mathcal{D}_{a_{2}}^{o_{2}} \subset \mathbb{R}^{H \times W \times 3}$, e.g., summer $\left(a_{2}\right)$ without occlusion $\left(o_{2}\right)$. As illustrated in Figure~\ref{fig:framework}, our framework consists of a shared place encoder $E^{p}$, a shared occlusion encoder $E^{o}$, two appearance encoders $\left\{E_{a_{1}}^{a}, E_{a_{2}}^{a}\right\}$, two generators $\left\{G_{a_{1}}, G_{a_{2}}\right\}$, four discriminators $\left\{D_{a_{1}}, D_{a_{2}}, D_{o_{1}}, D_{o_{2}}\right\}$, two for appearance and the others for occlusion, and a place discriminator $D_{p}$. Taking domain $\mathcal{D}_{a_{1}}^{o_{1}}$ as an example, the place encoder $E^{p}$ maps images onto a shared, domain-invariant place space $\left(E^{p}: \mathcal{D}_{a_{1}}^{o_{1}} \rightarrow \mathcal{P}\right)$, the occlusion encoder $E^{o}$ maps images onto a shared occlusion space $\left(E^{o}: \mathcal{D}_{a_{1}}^{o_{1}} \rightarrow \mathcal{O}\right)$, and the appearance encoder $E_{a_{1}}^{a}$ maps images onto a domain-specific appearance space $\left(E_{a_{1}}^{a}: \mathcal{D}_{a_{1}}^{o_{1}} \rightarrow \mathcal{A}_{a_{1}}\right)$. The generator $G_{a_{1}}$ synthesizes images conditioned on place, occlusion and appearance codes $\left(G_{a_{1}}:\left\{\mathcal{P}, \mathcal{O}, \mathcal{A}_{a_{1}}\right\} \rightarrow \mathcal{D}_{a_{1}}^{o_{1}} \right)$. The discriminator $D_{a_{1}}$ and $D_{o_{1}}$ aim to discriminate between real and translated images in the domain $\mathcal{D}_{a_{1}}$ and $\mathcal{D}_{o_{1}}$, respectively. In addition, the place discriminator $D_{p}$ is trained to distinguish the place codes extracted from two domains.

\subsection{Cross-cycle Consistency Constraint}
Thanks to the disentangled representations, we can perform image-to-image translation by combining a place code from a given image with an occlusion code and an appearance code from an image of the target domain. However, training with unpaired images is inherently ill-posed and requires additional constraints. Inspired by recent works~\cite{DRIT,DRIT_plus} that exploit the disentangled content and appearance codes for cyclic reconstruction, we propose a modified cross-cycle consistency loss to regularize training. 

Given unpaired images $x_{a_{1}}^{o_{1}}$ and $y_{a_{2}}^{o_{2}}$, we encode them into:
\begin{equation}\label{eq1}
\begin{aligned}
&\left\{z_{x}^{p}, z_{x}^{o}, z_{x}^{a}\right\}=\left\{E^{p}(x_{a_{1}}^{o_{1}}), E^{o}(x_{a_{1}}^{o_{1}}), E_{a_{1}}^{a}(x_{a_{1}}^{o_{1}})\right\}, \\
&\left\{z_{y}^{p}, z_{y}^{o}, z_{y}^{a}\right\}=\left\{E^{p}(y_{a_{2}}^{o_{2}}), E^{o}(y_{a_{2}}^{o_{2}}), E_{a_{2}}^{a}(y_{a_{2}}^{o_{2}})\right\}.
\end{aligned}
\end{equation}

We then perform the first translation by swapping the place representation to generate translated images $\left\{\bar{y}_{a_{1}}^{o_{1}}, \bar{x}_{a_{2}}^{o_{2}}\right\}$, where $\bar{y}_{a_{1}}^{o_{1}} \in \mathcal{D}_{a_{1}}^{o_{1}}$ and $\bar{x}_{a_{2}}^{o_{2}} \in \mathcal{D}_{a_{2}}^{o_{2}}$.
\begin{equation}\label{eq2}
\bar{y}_{a_{1}}^{o_{1}}=G_{a_{1}}\left(z_{y}^{p}, z_{x}^{o}, z_{x}^{a}\right), \quad \bar{x}_{a_{2}}^{o_{2}}=G_{a_{2}}\left(z_{x}^{p}, z_{y}^{o}, z_{y}^{a}\right).
\end{equation}

Next, we encode the translated images $\bar{y}_{a_{1}}^{o_{1}}$ and $\bar{x}_{a_{2}}^{o_{2}}$ into $\left\{z_{\bar{y}}^{p}, z_{\bar{y}}^{o}, z_{\bar{y}}^{a}\right\}$ and $\left\{z_{\bar{x}}^{p}, z_{\bar{x}}^{o}, z_{\bar{x}}^{a}\right\}$, and perform the second translation by swapping the place representation again to generate reconstructed images $\left\{\hat{x}_{a_{1}}^{o_{1}}, \hat{y}_{a_{2}}^{o_{2}}\right\}$.
\begin{equation}\label{eq3}
\begin{array}{r}
\hat{x}_{a_{1}}^{o_{1}}=G_{a_{1}}\left(z_{\bar{x}}^{p}, z_{\bar{y}}^{o}, z_{\bar{y}}^{a}\right), \quad \hat{y}_{a_{2}}^{o_{2}}=G_{a_{2}}\left(z_{\bar{y}}^{p}, z_{\bar{x}}^{o}, z_{\bar{x}}^{a}\right).
\end{array}
\end{equation}

Here we assume that the reconstructed images should be the same as the original images, and the cross-cycle consistency loss is formulated as:
\begin{equation}\label{eq4}
\begin{aligned}
\mathcal{L}_{1}^{\mathrm{cc}}&\left( G_{a_{1}}, G_{a_{2}}, E^{p}, E^{o}, E_{a_{1}}^{a}, E_{a_{2}}^{a} \right)= \\
&\mathbb{E}_{x_{a_{1}}^{o_{1}}, y_{a_{2}}^{o_{2}}}\left[\left\|\hat{x}_{a_{1}}^{o_{1}} -x_{a_{1}}^{o_{1}}\right\|_{1} + \left\|\hat{y}_{a_{2}}^{o_{2}} -y_{a_{2}}^{o_{2}}\right\|_{1}\right],
\end{aligned}
\end{equation}
where we have:
\begin{equation}\label{eq5}
\begin{array}{r}
\hat{x}_{a_{1}}^{o_{1}}= G_{a_{1}}\left( E^{p}\left( \bar{x}_{a_{2}}^{o_{2}} \right) ,E^{o}\left( \bar{y}_{a_{1}}^{o_{1}} \right), E_{a_{1}}^{a}\left( \bar{y}_{a_{1}}^{o_{1}} \right) \right), \\
\hat{y}_{a_{2}}^{o_{2}}= G_{a_{2}}\left( E^{p}\left( \bar{y}_{a_{1}}^{o_{1}} \right) ,E^{o}\left( \bar{x}_{a_{2}}^{o_{2}} \right), E_{a_{2}}^{a}\left( \bar{x}_{a_{2}}^{o_{2}} \right) \right).
\end{array}
\end{equation}

\subsection{Adversarial Constraint}
We employ domain adversarial losses $\mathcal{L}_{\mathrm{adv}}^{\mathrm{appearance}}$ and $\mathcal{L}_{\mathrm{adv}}^{\mathrm{occlusion}}$ in image level to encourage the appearance and the occlusion codes to carry appearance properties and occlusion content, respectively.
\begin{equation}\label{eq6}
\begin{aligned}
&\mathcal{L}_{\mathrm{adv}}^{\mathrm{appearance}}\left( G_{a_{1}}, G_{a_{2}}, E^{p}, E^{o}, E_{a_{1}}^{a}, E_{a_{2}}^{a}, D_{a_{1}}, D_{a_{2}} \right) = \\
& \mathbb{E}_{\bar{x}_{a_{2}}^{o_{2}}}\left[\log \left(1 - D_{a_{2}}\left(\bar{x}_{a_{2}}^{o_{2}}\right)\right)\right] +\mathbb{E}_{y_{a_{2}}^{o_{2}}}\left[\log D_{a_{2}}\left(y_{a_{2}}^{o_{2}}\right)\right] + \\
&\mathbb{E}_{\bar{y}_{a_{1}}^{o_{1}}}\left[\log \left(1 - D_{a_{1}}\left(\bar{y}_{a_{1}}^{o_{1}}\right)\right)\right] +\mathbb{E}_{x_{a_{1}}^{o_{1}}}\left[\log D_{a_{1}}\left(x_{a_{1}}^{o_{1}}\right)\right] \\
\end{aligned}
\end{equation}
where generators $G_{a_{1}}$ and $G_{a_{2}}$ are trained to fool discriminators $D_{a_{1}}$ and $D_{a_{2}}$ that try to distinguish between generated and real images adversely. Discriminators $D_{o_{1}}$ and $D_{o_{2}}$ and loss $\mathcal{L}_{\mathrm{adv}}^{\mathrm{occlusion}}$ are defined similarly.

On the other hand, we apply a place adversarial loss $\mathcal{L}_{\mathrm{adv}}^{\mathrm{place}}$ in feature level with a place discriminator $D_{p}$ that aims to distinguish the place features $z_{x}^{p}$ and $z_{y}^{p}$. The place encoder $E^{p}$ learns to encode indistinguishable codes to achieve disentanglement of domain-invariant place space $\mathcal{P}$.
\begin{equation}\label{equ7}
\begin{aligned}
\mathcal{L}_{\mathrm{adv}}^{\mathrm{place}}&\left(E^{p}, D_{p}\right)=\mathbb{E}_{x_{a_{1}}^{o_{1}}}\left[\log D_{p}\left(E^{p}(x_{a_{1}}^{o_{1}})\right)\right] \\
&+ \mathbb{E}_{y_{a_{2}}^{o_{2}}}\left[\log \left(1-D_{p}\left(E^{p}(y_{a_{2}}^{o_{2}})\right)\right)\right]
\end{aligned}
\end{equation}

\subsection{Geometry Constraint}
Since the place encoder and the occlusion encoder are symmetrical, the generator tends to ignore the place encoder and directly obtain all contents from the occlusion encoder to reconstruct the image, which results in the failure of feature disentanglement. To handle this issue,
we propose a geometry consistency loss $\mathcal{L}_{\mathrm{1}}^{\mathrm{gc}}$ and a cross-cycle geometry consistency loss $\mathcal{L}_{\mathrm{1}}^{\mathrm{cgc}}$ inspired by~\cite{gcgan}, which enables one-sided unsupervised domain mapping under the assumption that simple geometric transformations do not change the semantic structure of images.

As shown in Figure~\ref{fig:framework}, given a predefined geometric transformation function $f(\cdot)$, we can obtain counterpart images $x_{a_{1}}^{{o_{1}}^{\prime}}$ and $y_{a_{2}}^{{o_{2}}^{\prime}}$ by applying $f(\cdot)$ to $x_{a_{1}}^{o_{1}}$ and $y_{a_{2}}^{o_{2}}$, respectively. We learn additional translations as:
\begin{equation}\label{eq8}
\begin{array}{r}
\hat{y}_{a_{2}}^{{o_{2}}^{\prime}}= G_{a_{2}}( E^{p}( \bar{y}_{a_{1}}^{{o_{1}}^{\prime}}),z_{\bar{x}}^{o}, z_{\bar{x}}^{a}), \\
\bar{y}_{a_{1}}^{{o_{1}}^{\prime}}= G_{a_{1}}( E^{p}( y_{a_{2}}^{{o_{2}}^{\prime}}), z_{x}^{o}, z_{x}^{a}), \\
\bar{x}_{a_{2}}^{{o_{2}}^{\prime}}= G_{a_{2}}( E^{p}( x_{a_{1}}^{{o_{1}}^{\prime}}), z_{y}^{o}, z_{y}^{a}).
\end{array}
\end{equation}

Since we design the appearance code $z_{y}^{a}$ as a global representation that is invariant to geometric transformation and the occlusion code $z_{y}^{o}$ contains no occlusion information when domain $\mathcal{D}_{o_{2}}$ is \textit{without occlusion}, we can assume that $\bar{x}_{a_{2}}^{{o_{2}}^{\prime}}$ and $\bar{x}_{a_{2}}^{o_{2}}$ should keep the same geometric transformation with the one between $x_{a_{1}}^{{o_{1}}^{\prime}}$ and $x_{a_{1}}^{o_{1}}$, i.e.,$f\left(\bar{x}_{a_{2}}^{o_{2}}\right) \approx \bar{x}_{a_{2}}^{{o_{2}}^{\prime}}$, where $f\left(x_{a_{1}}^{o_{1}}\right) = x_{a_{1}}^{{o_{1}}^{\prime}}$. To enforce this constraint, we formulate the geometry consistency loss as:
\begin{equation}\label{eq9}
\begin{aligned}
&\mathcal{L}_{\mathrm{1}}^{\mathrm{gc}}\left(G_{a_{2}}, E^{p}, E^{o}, E_{a_{2}}^{a} \right) = \\
&\mathbb{E}_{x_{a_{1}}^{o_{1}}, y_{a_{2}}^{o_{2}}}\left[\left\|\bar{x}_{a_{2}}^{o_{2}}-f^{-1}(\bar{x}_{a_{2}}^{{o_{2}}^{\prime}})\right\|_{1} + \left\|\bar{x}_{a_{2}}^{{o_{2}}^{\prime}}-f(\bar{x}_{a_{2}}^{o_{2}})\right\|_{1}\right].
\end{aligned}
\end{equation}

In this paper, we take $90^{\circ}$ clockwise rotation as a geometric transformation example.
Similar to the geometry consistency constraint,
we further assume that the reconstructed image $\hat{y}_{a_{2}}^{{o_{2}}^{\prime}}$ should be the same as $y_{a_{2}}^{{o_{2}}^{\prime}}$, 
and express this cross-cycle geometry consistency loss as: 
\begin{equation}\label{eq10}
\begin{aligned}
\mathcal{L}_{\mathrm{1}}^{\mathrm{cgc}}&\left( G_{a_{1}}, G_{a_{2}}, E^{p}, E^{o}, E_{a_{1}}^{a}, E_{a_{2}}^{a} \right) = \\
&\mathbb{E}_{x_{a_{1}}^{o_{1}}, y_{a_{2}}^{o_{2}}}\left[\left\|\hat{y}_{a_{2}}^{{o_{2}}^{\prime}}-y_{a_{2}}^{{o_{2}}^{\prime}}\right\|_{1}\right].
\end{aligned}
\end{equation}

   \begin{figure}[t]
      \includegraphics[width=1\linewidth]{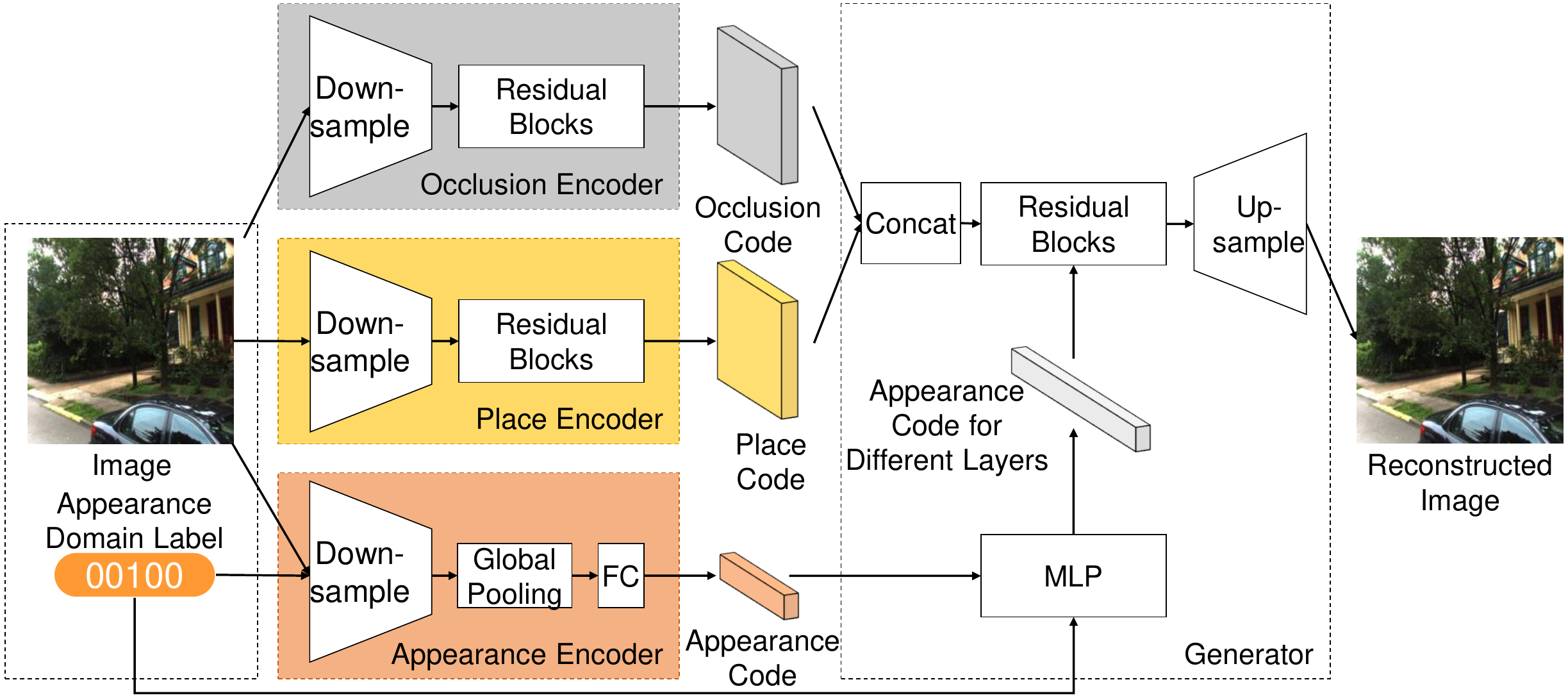}
      \caption{Architecture of PROCA. 
      Different encoders encode the image to respective codes. 
      The generator uses an MLP to produce different appearance codes for different layers. We concatenate the place and occlusion code and then process them by residual blocks under the guidance of the appearance code. Finally, the reconstructed image can be generated conditioned on the three codes.
      }
      \vspace{-0.5cm}
      \label{fig:network}
   \end{figure}

\subsection{Auxiliary Constraints}
In addition to the constraints above, we employ four other loss functions to facilitate network training, as illustrated in Figure~\ref{fig:auxiliary_constraint}.

\begin{table*}[t]
\caption{Comparison Under Different Conditions on CMU-Seasons Dataset}
\centering
\tiny
\resizebox{\linewidth}{!}{
\begin{tabular}{c|c|c|c|c|c}
\hline
\multirow{3}{*}{Methods}& Overcast$(\%)$                                                                    & Sunny$(\%)$                                                                   & Low Sun$(\%)$                                                                 & Cloudy$(\%)$                                                                  & Snow$(\%)$                                                                    \\
                         & \begin{tabular}[c]{@{}c@{}}0.25m / 0.5m / 5m\\ 2\degree\quad/ 5\degree\quad / 10\degree\quad \end{tabular} & \begin{tabular}[c]{@{}c@{}}0.25m / 0.5m / 5m\\ 2\degree\quad/ 5\degree\quad / 10\degree\quad \end{tabular} & \begin{tabular}[c]{@{}c@{}}0.25m / 0.5m / 5m\\ 2\degree\quad/ 5\degree\quad / 10\degree\quad \end{tabular} & \begin{tabular}[c]{@{}c@{}}0.25m / 0.5m / 5m\\ 2\degree\quad/ 5\degree\quad / 10\degree\quad \end{tabular} & \begin{tabular}[c]{@{}c@{}}0.25m / 0.5m / 5m\\ 2\degree\quad/ 5\degree\quad / 10\degree\quad \end{tabular} \\ \hline
FAB-MAP~\cite{FAB-MAP}                  & 0.9 / 2.7 / 17.0                                                            & 1.0 / 2.5 / 15.2                                                        & 2.0 / 4.6 / 20.8                                                        & 1.8 / 4.1 / 20.1                                                        & 2.2 / 4.8 / 22.4                                                        \\
NetVLAD~\cite{netvlad}                  & 10.9 / 27.0 / 82.7                                                          & 10.5 / 25.9 / 79.2                                                      & 10.1 / 25.7 / 77.7                                                      & 13.0 / 30.5 / 82.9                                                      & 10.2 / 25.3 / 75.5                                                      \\
DenseVLAD~\cite{DenseVLAD}                & 15.1 / 35.2 / 85.2                                                          & 13.2 / 31.3 / 81.4                                                      & 15.1 / 36.9 / \textcolor{second}{86.0}                                                      & 18.4 / 41.8 / \textcolor{second}{89.0}                                                      & 17.4 / 41.3 / 87.2                                                      \\
DIFL-FCL~\cite{DIFLFCL}                 & 15.9 / 36.9 / 83.1                                                          & 14.1 / 32.7 / 78.7                                                      & 13.9 / 34.1 / 79.2                                                      & 16.4 / 37.6 / 84.8                                                      & 13.6 / 33.4 / 70.1                                                      \\
DISAM~\cite{DISAM}                    & 18.0 / 39.6 / 85.3                                                          & 15.2 / 33.9 / 80.9                                                      & 15.8 / 37.3 / 82.3                                                      & 18.6 / 40.5 / 87.6                                                      & 15.7 / 37.3 / 76.3                                                      \\ \hline
PROCA-O                     & 12.9 / 31.5 / 83.1                                                          & 11.4 / 27.1 / 79.5                                                      & 11.7 / 29.6 / 81.2                                                      & 15.5 / 32.9 / 83.4                                                      & 10.8 / 27.2 / 76.5                                                      \\
PROCA-A                     & \textcolor{second}{18.4} / \textcolor{second}{40.5} / \textcolor{second}{87.6}                                                          & \textcolor{second}{16.7} / \textcolor{second}{35.9} / \textcolor{second}{81.5}                                                      & \textcolor{second}{17.3} / \textcolor{second}{40.6} / 84.6                                                      & \textcolor{second}{19.7} / \textcolor{second}{42.4} / 88.3                                                      & \textcolor{second}{18.1} / \textcolor{second}{43.8} / \textcolor{second}{87.8}                                                      \\
PROCA                     & \textcolor{best}{19.5} / \textcolor{best}{43.9} / \textcolor{best}{88.4}                                                          & \textcolor{best}{17.2} / \textcolor{best}{38.9} / \textcolor{best}{82.9}                                                      & \textcolor{best}{17.6} / \textcolor{best}{42.1} / \textcolor{best}{87.7}                                                      & \textcolor{best}{20.0} / \textcolor{best}{44.4} / \textcolor{best}{90.4}                                                      & \textcolor{best}{18.3} / \textcolor{best}{44.3} / \textcolor{best}{89.6}                                                      \\ \hline
\end{tabular}
}
\vspace{-0.5cm}
\label{tab:overcast}
\end{table*}

\begin{table}[t]
\caption{Comparison Under Different Areas}
\centering
\Huge
\resizebox{\linewidth}{!}{
\begin{tabular}{c|c|c|c}
\hline
\multirow{3}{*}{Methods} & Urban$(\%)$                                                                           & Suburban$(\%)$                                                                        & Park$(\%)$                                                                            \\
                         & \begin{tabular}[c]{@{}c@{}}0.25m / 0.5m / 5m\\ 2\degree\quad/ 5\degree\quad / 10\degree\quad \end{tabular} & \begin{tabular}[c]{@{}c@{}}0.25m / 0.5m / 5m\\ 2\degree\quad/ 5\degree\quad / 10\degree\quad \end{tabular} & \begin{tabular}[c]{@{}c@{}}0.25m / 0.5m / 5m\\ 2\degree\quad/ 5\degree\quad / 10\degree\quad \end{tabular} \\ \hline
FAB-MAP                  & 2.7 / 6.4 / 27.3                                                                & 0.5 / 1.5 / 13.6                                                                & 0.8 / 1.7 / 11.5                                                                \\ 
NetVLAD                  & 17.4 / 40.3 / 93.2                                                              & 7.6 / 21.0 / 80.5                                                               & 5.6 / 15.7 / 65.8                                                               \\
DenseVLAD                & 22.2 / \textcolor{second}{48.6} / \textcolor{best}{92.8}                                                              & 9.8 / 26.6 / \textcolor{best}{85.2}                                                               & 10.3 / 27.1 / 77.0                                                              \\
DIFL-FCL                 & 20.2 / 44.7 / 87.8                                                              & 9.7 / 25.0 / 73.7                                                               & 11.4 / 28.9 / 76.4                                                              \\
DISAM                    & \textcolor{second}{22.6} / 47.3 / 89.1                                                              & \textcolor{second}{11.1} / \textcolor{second}{27.5} / 77.6                                                              & \textcolor{second}{12.6} / \textcolor{second}{31.3} / \textcolor{second}{80.0}                                                              \\ \hline
PROCA                    & \textcolor{best}{24.9} / \textcolor{best}{53.4} / \textcolor{second}{92.7}                                                              & \textcolor{best}{12.0} / \textcolor{best}{30.4} / \textcolor{second}{83.3}                                                              & \textcolor{best}{14.1} / \textcolor{best}{35.2} / \textcolor{best}{82.6}                                                              \\ \hline
\end{tabular}
}
\vspace{-0.5cm}
\label{tab:urban}
\end{table}

\textbf{Self-reconstruction loss.} In order to facilitate the training, we apply a self-reconstruction loss $\mathcal{L}_{\mathrm{1}}^{\mathrm{recon}}$. With encoded features $\left\{z_{x}^{p}, z_{x}^{o}, z_{x}^{a}\right\}$ and $\left\{z_{y}^{p}, z_{y}^{o}, z_{y}^{a}\right\}$, the generators $G_{a_{1}}$ and $G_{a_{2}}$ should be able to generate them back to the original images. That is, $\tilde{x}_{a_{1}}^{o_{1}}=G_{a_{1}}\left(E^{p}(x_{a_{1}}^{o_{1}}), E^{o}(x_{a_{1}}^{o_{1}}), E_{a_{1}}^{a}(x_{a_{1}}^{o_{1}})\right)$ and $\tilde{y}_{a_{2}}^{o_{2}}=G_{a_{2}}\left(E^{p}(y_{a_{2}}^{o_{2}}), E^{o}(y_{a_{2}}^{o_{2}}), E_{a_{2}}^{a}(y_{a_{2}}^{o_{2}})\right)$.

\textbf{KL loss.} To further disentangle the appearance code, we use KL loss $\mathcal{L}_{\mathrm{KL}}$ to encourage the appearance representation to be as close to a prior Gaussian distribution as possible, similar to~\cite{DRIT_plus}. The loss is  $\mathcal{L}_{\mathrm{KL}}=\mathbb{E}\left[D_{\mathrm{KL}}\left(\left(z_{a}\right) \| N(0,1)\right)\right]$, where $D_{\mathrm{KL}}(p \| q)=-\int p(z) \log \frac{p(z)}{q(z)} \mathrm{d}z$.

\textbf{Appearance latent regression loss.} To encourage invertible mapping between the image space and the appearance space, 
we apply an appearance latent regression loss $\mathcal{L}_{\mathrm{1}}^{\mathrm{appearance}}$,
drawing a latent code $z$ from the prior Gaussian distribution as the appearance code and attempting to reconstruct it with $\hat{z}=E_{a_{2}}^{a}\left(G_{{a_{2}}}\left(E^{p}(x_{a_{1}}^{o_{1}}), E^{o}(x_{a_{1}}^{o_{1}}), z\right)\right)$ and $\hat{z}=E_{a_{1}}^{a}\left(G_{{a_{1}}}\left(E^{p}(y_{a_{2}}^{o_{2}}), E^{o}(y_{a_{2}}^{o_{2}}), z\right)\right)$.

\textbf{Place latent regression loss.} Moreover, to further 
disentangle the place code 
and enhance the training efficiency, 
we propose a place latent regression loss $\mathcal{L}_{\mathrm{1}}^{\mathrm{place}}$ to enforce that the place code of original images in domain $\mathcal{D}_{o_{1}}$ (with occlusion) should be the same as the place code of transformed images in domain $\mathcal{D}_{o_{2}}$ (without occlusion). That means $E^{p}(x_{a_{1}}^{o_{1}})=E^{p}\left(G_{{a_{2}}}\left(E^{p}(x_{a_{1}}^{o_{1}}), E^{o}(y_{a_{2}}^{o_{2}}), E_{a_2}^{a}(y_{a_{2}}^{o_{2}})\right)\right)$, and the other side $E^{p}(y_{a_{2}}^{o_{2}})=E^{p}\left(G_{{a_{1}}}\left(E^{p}(y_{a_{2}}^{o_{2}}), E^{o}(x_{a_{1}}^{o_{1}}), E_{a_1}^{a}(x_{a_{1}}^{o_{1}})\right)\right)$ is defined in a similar manner.

\textbf{Total loss.} To achieve disentanglement of the place code, occlusion code and appearance code, we combine the aforementioned constraints and jointly train the encoders, generators, and discriminators to optimize the full objective $\mathcal{L}_{\mathrm{PROCA}}\left( G_{a_{1}}, G_{a_{2}}, E^{p}, E^{o}, E_{a_{1}}^{a}, E_{a_{2}}^{a}, D_{a_{1}}, D_{a_{2}} \right)$:
\begin{equation}\label{eq11}
\begin{aligned}
&\min _{G_{a_{1}}, G_{a_{2}}, E^{p}, E^{o}, E_{a_{1}}^{a}, E_{a_{2}}^{a}} \max _{D_{a_{1}}, D_{a_{2}}} \mathcal{L}_{\mathrm{PROCA}}=\\  &\lambda^{cc}_{1}\mathcal{L}_{1}^{\mathrm{cc}}
+\lambda_{\mathrm{1}}^{\mathrm{gc}}\mathcal{L}_{\mathrm{1}}^{\mathrm{gc}} + \lambda_{\mathrm{1}}^{\mathrm{cgc}}\mathcal{L}_{\mathrm{1}}^{\mathrm{cgc}} + \lambda_{\mathrm{1}}^{\mathrm{recon}}\mathcal{L}_{\mathrm{1}}^{\mathrm{recon}} + \\
& \lambda_{\mathrm{adv}}^{\mathrm{app}}\mathcal{L}_{\mathrm{adv}}^{\mathrm{appearance}} +
\lambda_{\mathrm{adv}}^{\mathrm{occ}}\mathcal{L}_{\mathrm{adv}}^{\mathrm{occlusion}} + 
\lambda_{\mathrm{adv}}^{\mathrm{place}}\mathcal{L}_{\mathrm{adv}}^{\mathrm{place}} +\\
& \lambda_{\mathrm{1}}^{\mathrm{app}}\mathcal{L}_{\mathrm{1}}^{\mathrm{appearance}} + \lambda_{\mathrm{1}}^{\mathrm{place}}\mathcal{L}_{\mathrm{1}}^{\mathrm{place}} +\lambda_{\mathrm{KL}}\mathcal{L}_{\mathrm{KL}}\\
\end{aligned}
\end{equation}
where $\lambda$s are the trade-off hyper-parameters to control the contribution of each loss term. Tuning the hyper-parameters may give preferable results for specific tasks.

\subsection{Multi-Domain Extension}
However, the conditions are often more than two domains (e.g., winter with occlusion, summer without occlusion, spring with occlusion and autumn without occlusion). In order to train among multiple domains in one model, we propose the multi-domain extension of PROCA.

Given two images $(x_{a_{m}}^{o_{h}}, y_{a_{n}}^{o_{v}})$ randomly sampled from $k \times 2$ domains $\left\{\mathcal{D}_{a_{i}}^{o_{j}}\right\}_{i=1 \sim k, j=1 \sim 2}$, we can construct their one-hot appearance domain codes $\left(z_{m}^{d}, z_{n}^{d}\right)$, where $\left(x_{a_{m}}^{o_{h}} \in \mathcal{D}_{a_{m}}^{o_{h}}, y_{a_{n}}^{o_{v}} \in \mathcal{D}_{a_{n}}^{o_{v}}, Z^{d} \subset \mathbb{R}^{k}\right)$. We encode the images onto a shared place space $\mathcal{P}$, a shared occlusion space $\mathcal{O}$, and domain-specific appearance spaces $\left\{\mathcal{A}_{i}\right\}_{i=1 \sim k}$:
\begin{equation}\label{eq12}
\begin{aligned}
&\left\{z_{x}^{p}, z_{x}^{o}, z_{x}^{a}\right\}=\left\{E^{p}(x_{a_{m}}^{o_{h}}), E^{o}(x_{a_{m}}^{o_{h}}), E^{a}(x_{a_{m}}^{o_{h}}, z_{m}^{d})\right\},\\
&\left\{z_{y}^{p}, z_{y}^{o}, z_{y}^{a}\right\}=\left\{E^{p}(y_{a_{n}}^{o_{v}}), E^{o}(y_{a_{n}}^{o_{v}}), E^{a}(y_{a_{n}}^{o_{v}}, z_{n}^{d})\right\}.
\end{aligned}
\end{equation}
We then perform the cross-cycle translation and self-reconstruction similar to the dual-domain translation.
\begin{equation}\label{eq13}
\begin{aligned}
&\bar{y}_{a_{m}}^{o_{h}}=G\left(z_{y}^{p}, z_{x}^{o}, z_{x}^{a},  z_{m}^{d}\right), \bar{x}_{a_{n}}^{o_{v}}=G\left(z_{x}^{p}, z_{y}^{o}, z_{y}^{a}, z_{n}^{d}\right). \\
&\hat{x}_{a_{m}}^{o_{h}}=G\left(z_{\bar{x}}^{p}, z_{\bar{y}}^{o}, z_{\bar{y}}^{a}, z_{m}^{d}\right), \hat{y}_{a_{n}}^{o_{v}}=G\left(z_{\bar{y}}^{p}, z_{\bar{x}}^{o}, z_{\bar{x}}^{a}, z_{n}^{d}\right), \\
&\tilde{x}_{a_{m}}^{o_{h}}=G\left(z_{x}^{p}, z_{x}^{o}, z_{x}^{a},  z_{m}^{d}\right), \tilde{y}_{a_{n}}^{o_{v}}=G\left(z_{y}^{p}, z_{y}^{o}, z_{y}^{a}, z_{n}^{d}\right).
\end{aligned}
\end{equation}
In order to train with multiple domains without increasing discriminators, we leverage discriminator $D$ as an auxiliary domain classifier. That is, the discriminator $D$ not only aims to discriminate between real and translated images, but also performs domain classification.
\begin{equation}\label{eq14}
\begin{aligned}
&\mathcal{L}_{\mathrm{cls}}^{\mathrm{appearance}}\left( D \right)= \\
&\mathbb{E}_{x_{a_{m}}^{o_{h}}, y_{a_{n}}^{o_{v}}, z_{m}^{d}, z_{n}^{d}}\left[\log \frac{1}{D(z_{m}^{d} \mid x_{a_{m}}^{o_{h}}) D(z_{n}^{d} \mid y_{a_{n}}^{o_{v}})}\right]\\
&\mathcal{L}_{\mathrm{cls}}^{\mathrm{appearance}}\left( G_{a_{1}}, G_{a_{2}}, E^{p}, E^{o}, E_{a_{1}}^{a}, E_{a_{2}}^{a} \right) = \\
&\mathbb{E}_{x_{a_{m}}^{o_{h}}, y_{a_{n}}^{o_{v}}, z_{m}^{d}, z_{n}^{d}}\left[\log \frac{1}{D(z_{n}^{d} \mid \bar{x}_{a_{n}}^{o_{v}}) D(z_{m}^{d} \mid \bar{y}_{a_{m}}^{o_{h}})} \right].
\end{aligned}
\end{equation}
We incorporate this term into the full objective for the multi-domain extension of PROCA:
\begin{equation}\label{eq15}
\mathcal{L}_{\mathrm{PROCA-MD}}= \mathcal{L}_{\mathrm{PROCA}} + \mathcal{L}_{\mathrm{cls}}^{\mathrm{appearance}}
\end{equation}

\begin{table}[t]
\caption{Comparison Under Different Foliage Categories}
\centering
\Huge
\resizebox{\linewidth}{!}{
\begin{tabular}{c|c|c|c}
\hline
\multirow{3}{*}{Methods} & Foliage$(\%)$                                                                     & Mixed Foliage$(\%)$                                                           & No Foliage$(\%)$                                                              \\
                         & \begin{tabular}[c]{@{}c@{}}0.25m / 0.5m / 5m\\ 2\degree\quad/ 5\degree\quad / 10\degree\quad \end{tabular} & \begin{tabular}[c]{@{}c@{}}0.25m / 0.5m / 5m\\ 2\degree\quad/ 5\degree\quad / 10\degree\quad \end{tabular} & \begin{tabular}[c]{@{}c@{}}0.25m / 0.5m / 5m\\ 2\degree\quad/ 5\degree\quad / 10\degree\quad \end{tabular} \\ \hline
FAB-MAP                  & 1.1 / 2.7 / 16.5                                                            & 1.0 / 2.5 / 14.7                                                        & 3.6 / 7.9 / 30.7                                                        \\
NetVLAD                  & 10.4 / 26.0 / 80.1                                                          & 11.0 / 26.7 / 78.4                                                      & 11.8 / 29.2 / 82.0                                                      \\
DenseVLAD                & 13.2 / 31.6 / \textcolor{second}{82.3}                                                          & 16.2 / 38.1 / 85.4                                                      & \textcolor{best}{17.8} / \textcolor{best}{42.1} / \textcolor{best}{91.3}                                                      \\
DIFL-FCL                 & 13.9 / 32.7 / 79.3                                                          & 16.6 / 38.6 / 84.4                                                      & 12.9 / 32.2 / 76.5                                                      \\ 
DISAM                    & \textcolor{second}{15.2} / \textcolor{second}{34.1} / 81.6                                                          & \textcolor{second}{18.7} / \textcolor{second}{41.7} / \textcolor{second}{86.9}                                                      & 15.2 / 36.1 / 80.3                                                      \\\hline
PROCA                     & \textcolor{best}{17.1} / \textcolor{best}{39.0} / \textcolor{best}{84.0}                                                          & \textcolor{best}{20.2} / \textcolor{best}{45.8} / \textcolor{best}{89.7}                                                      & \textcolor{second}{16.9} / \textcolor{second}{41.0} / \textcolor{second}{88.5}                                                      \\ \hline
\end{tabular}
}
\label{tab:foliage}
\vspace{-0.5cm}
\end{table}

\section{Experiments}
\subsection{Implementation Details}
Figure~\ref{fig:network} shows the architecture of PROCA. The place encoder $E^{p}$ and occlusion encoder $E^{o}$ consist of convolution layers followed by residual blocks. The appearance encoder $E^{a}$ consists of convolution layers followed by fully-connected layers. The generator $G$ consists of residual blocks followed by fractionally strided convolution layers, and also uses a multilayer perceptron to produce different appearance codes for different layers from the original appearance code and domain label. 
We use a concatenation operation 
to fuse the occlusion code and the place code. 
We follow the procedure in DRIT~\cite{DRIT} for training the model. The images are resized to $256 \times 256$ and cropped to $216 \times 216$ sizes randomly while training, and we directly resize the images to $216 \times 216$ while testing. Dimension of the encoded features code is flatted after the output of place encoder with a shape of $256 \times 54 \times 54$. 
More details can be found at https://github.com/rover-xingyu/PROCA.

\begin{figure}[t]
    \centering
    \vspace{-0.3cm}
	  \subfloat[All]{
       \includegraphics[height=0.11\textwidth]{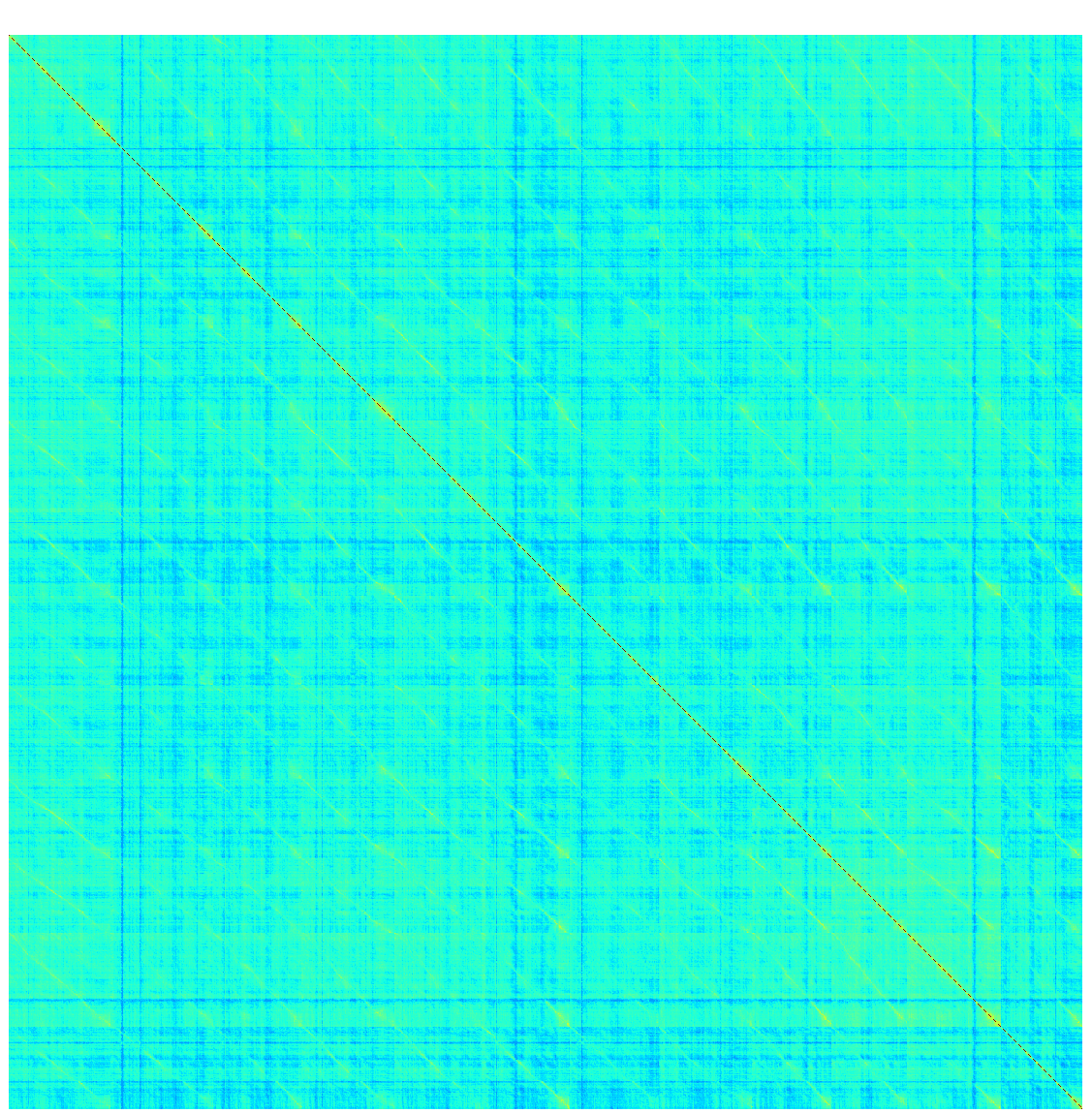}}
    \label{fig:s1}\hfill
	  \subfloat[Appearance]{
        \includegraphics[height=0.11\textwidth]{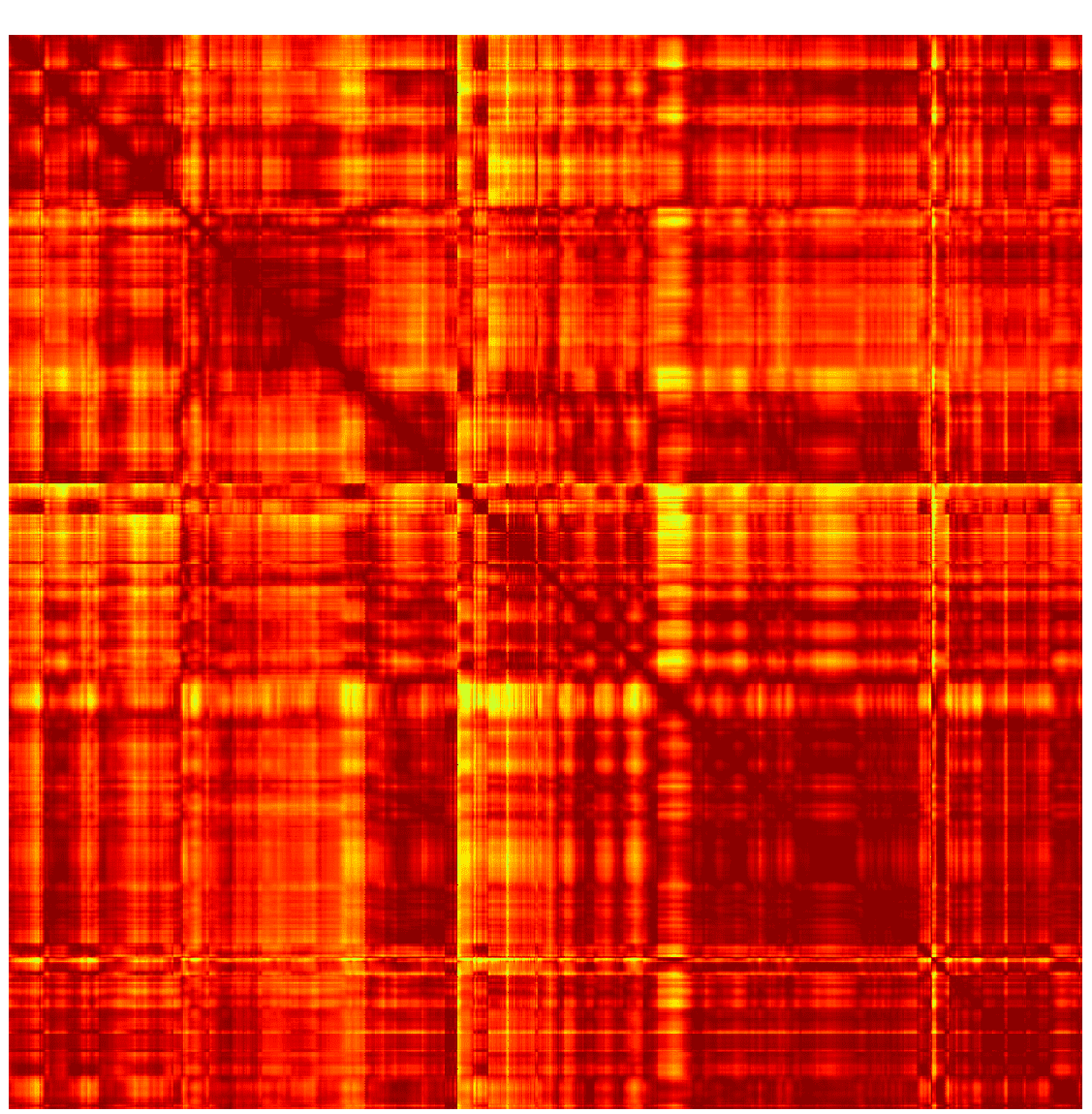}}
    \label{fig:s2}\hfill
	  \subfloat[Occlusion]{
        \includegraphics[height=0.11\textwidth]{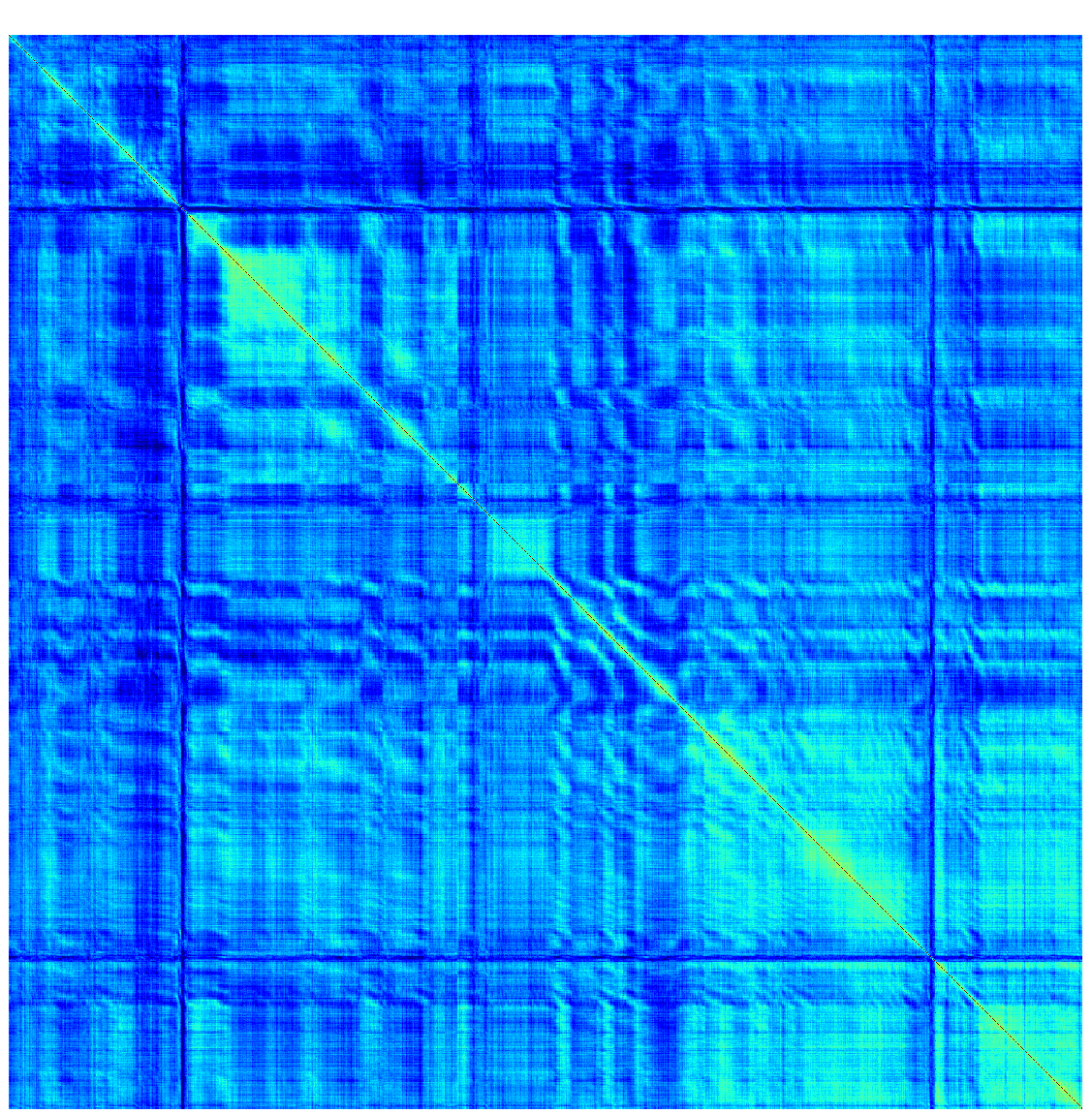}}
    \label{fig:s3}\hfill
	  \subfloat[Place]{
        \includegraphics[height=0.11\textwidth]{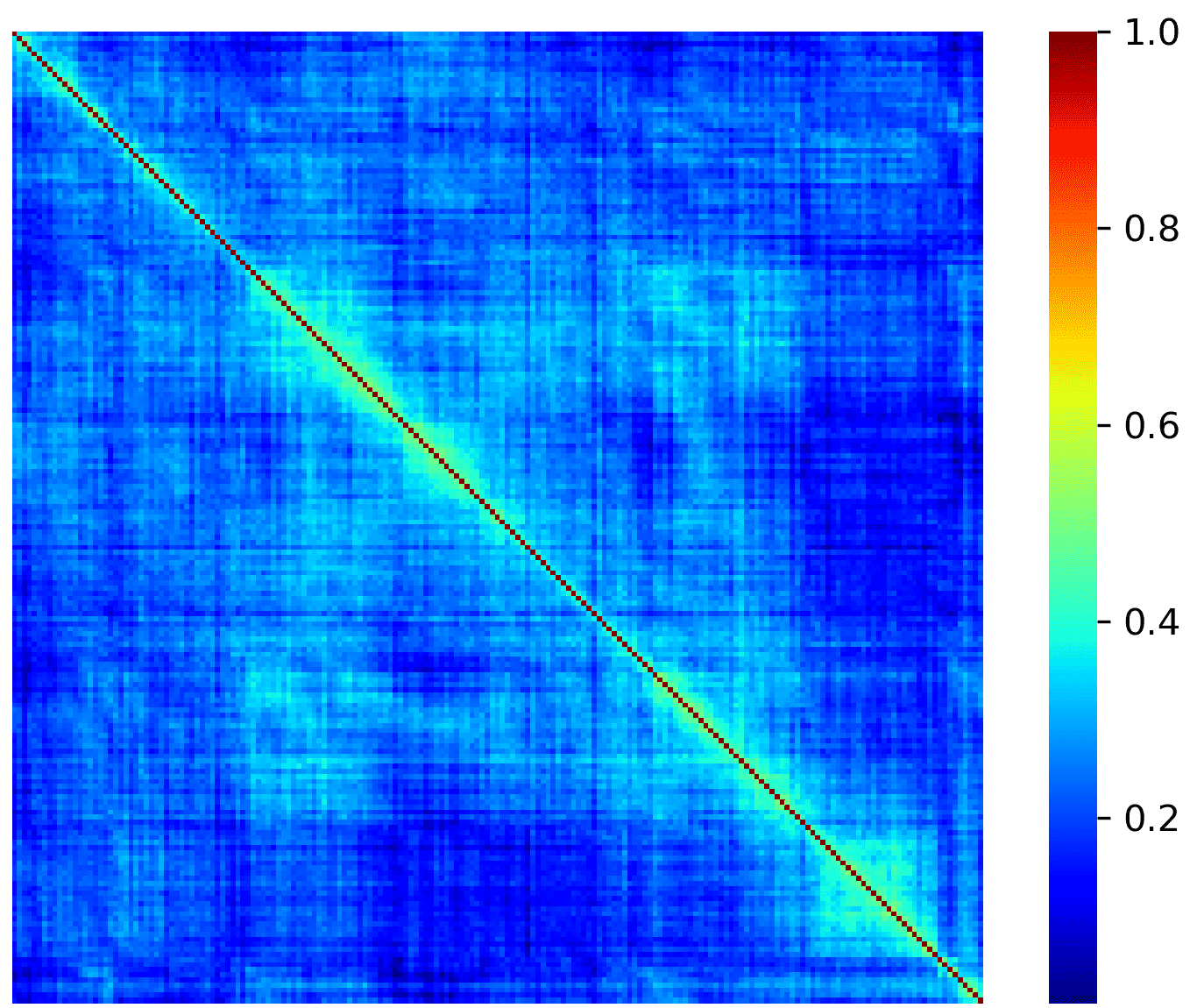}}
     \label{fig:s4} 
	\caption{Similarity matrices of (a) ``All'' codes encoded by place encoder of PROCA but without disentanglement training, and (b) appearance, (c) occlusion, (d) place codes of PROCA.}
	\vspace{-0.5cm}
	  \label{fig:similarity} 
\end{figure}

\subsection{Experimental Setup}
The experiments are conducted on the CMU-Seasons dataset~\cite{Benchmarking,cmudata}. It was recorded over the course of a year by having a vehicle with cameras drive on a 9 kilometers long route. The dataset is challenging due to the variance of environmental conditions as a result of changing seasons, illumination, weather, foliage, and occlusion. It is benchmarked in~\cite{Benchmarking}, which gives a clear category and division of the original dataset, together with the groud-truth of camera pose per reference database image. There are three areas in the CMU-Seasons dataset: 31250 images for the urban, 13736 images for the suburban and 30349 images for the parks. Additionally, there are one reference and eleven query conditions for each area. The reference database is recorded under the condition of sunny with no foliage, while the query image can be chosen from sunny, cloudy, overcast, snow, etc. intersected with dense, mixed or no foliage. We further label the images with occlusion and without occlusion depending on if there are dynamic objects in the images. In order to verify the generalization ability of PROCA, we sample 9055 images with occlusion and 9068 images without occlusion from the urban part as the training set, and evaluate on the untrained suburban and park parts.

\subsection{Evaluation}

\noindent\textbf{Baselines.}
We evaluate our proposed method against FAB-MAP~\cite{FAB-MAP}, NetVLAD~\cite{netvlad}, DenseVLAD~\cite{DenseVLAD}, DIFL-FCL~\cite{DIFLFCL}, and DISAM~\cite{DISAM}), and two ablations of PROCA: PROCA-A and PROCA-O. PROCA-A (appearance) builds upon our full model by eliminating the anti-occlusion components, and PROCA-O (occlusion) removes the appearance handling components from the full model. PROCA represents the complete model of our method.

\noindent\textbf{Comparisons.}
Following the evaluation protocol introduced in~\cite{Benchmarking}, we report the percentage of query images correctly localized within three different 6-DOF pose error tolerance thresholds: $\left(0.25 m, 2^{\circ}\right)$,  $\left(0.5 m, 5^{\circ}\right)$ and $\left(5 m, 10^{\circ}\right)$.

Table~\ref{tab:overcast} shows results under different conditions for all ablations and baselines, where the results of baselines come form~\cite{Benchmarking}. 
Our proposed method achieves higher accuracy than all baselines in all conditions. 

Results under different areas are summarized in Table \ref{tab:urban}, where PROCA outperforms baselines for high- and medium-precision localization in all areas, which indicates our powerful generalization ability because the model is only trained in urban areas. 

Though DISAM adopts domain-invariant feature learning methods to handle appearance variations, the medium-precision localization in the urban areas of DISAM is suffered from numerous dynamic objects.

We further compare the performance on different foliage categories in Table \ref{tab:foliage}, from which we can see that our results are better than baselines under foliage and mixed foliage. DenseVLAD performs better than ours on no foliage query images because the reference database is also captured under no foliage, with no need to consider the domain gap.

\begin{figure}[t]
  \includegraphics[width=1\linewidth]{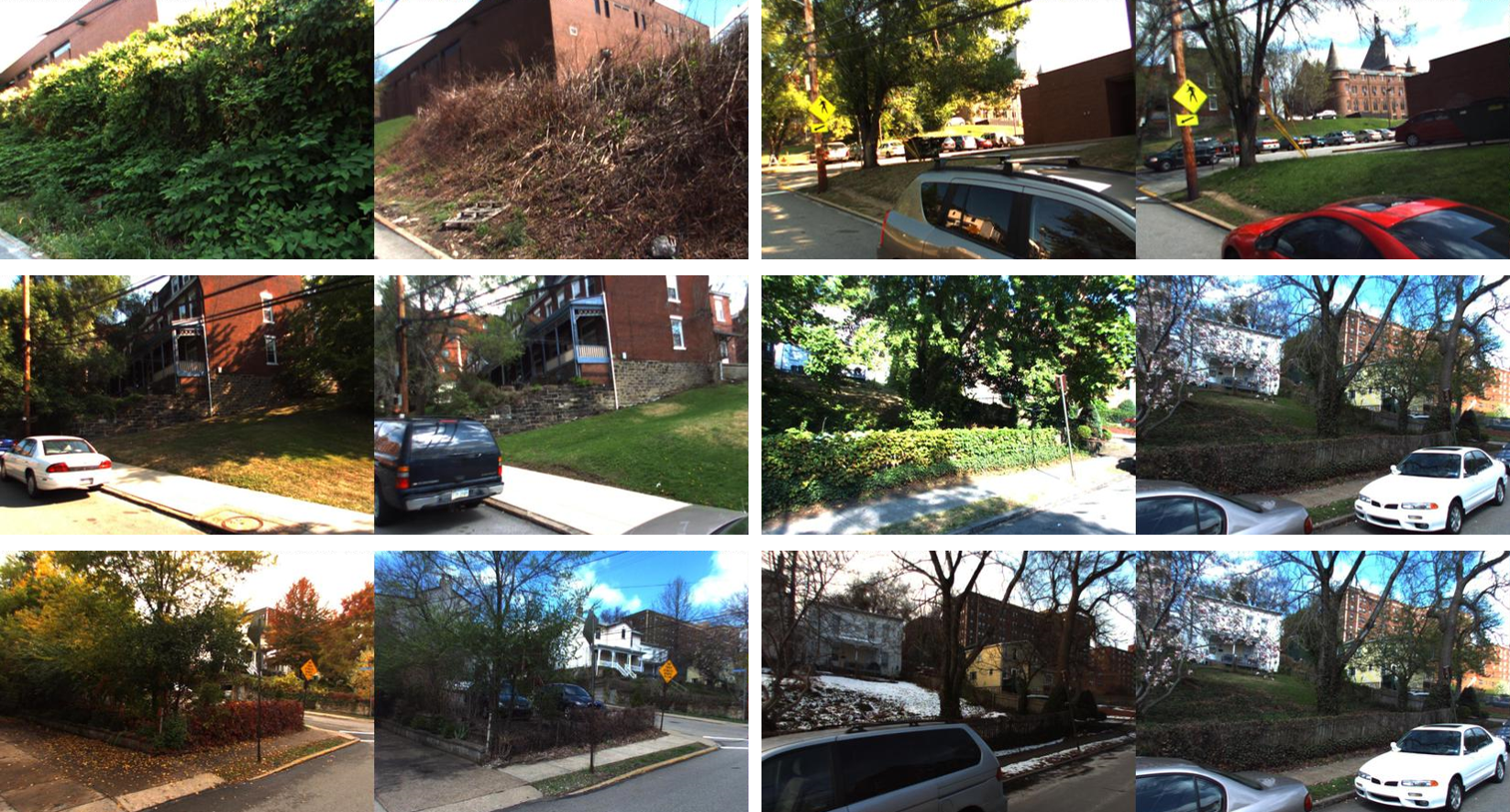}
  \caption{Examples of matching pairs of PROCA. For each pair of images, the left is the query image under diverse conditions, while the right is the retrieved database image.}
  \label{fig:example}
  \vspace{-0.5cm}
\end{figure}

\noindent\textbf{Qualitative analysis.} Fig.~\ref{fig:similarity} illustrates similarity matrices for an intuitive visual experience of the effective feature disentanglement of PROCA. We extract the codes from a sequence of images and calculate its similarity matrix, where the ground-truth similarity matrix is diagonal. It can be seen that the disentangled place code is significantly distinctive, thus preferable in place recognition. On the contrary, the appearance and occlusion codes are hard to distinguish among different places and thus should be decomposed from ``All'' codes. Examples of image pairs matched by PROCA are shown in Fig.~\ref{fig:example}. Leveraging disentangled representations, we could perform impressive place recognition ability under extreme occlusion or changing appearance.

\noindent\textbf{Limitations.} Based on the fact that PROCA is trained with the auxiliary supervision of unsupervised image-to-image translation, our network is more robust to domain changes but less to perspective shifts, which gives a lower performance than DenseVLAD when there is no domain gap. Hence developing multiple view geometry constraints is essential to enable the perspective robustness of PROCA.

\section{Conclusions}
Place recognition has grown in prominence and can be utilized in various applications, including VR/AR, mobile robots and autonomous vehicles. While conventional methods work effectively with perspective shifts, it is hard to tackle the domain gap. To overcome this challenging problem, we present a novel framework for place recognition in an unsupervised manner, disentangling the place latent space from occlusion and appearance space. Extensive experiments have demonstrated the effectiveness of our method.

{\small{
\bibliographystyle{IEEEtran}
\bibliography{cite}}

\begin{thebibliography}{10}
\providecommand{\url}[1]{#1}
\csname url@samestyle\endcsname
\providecommand{\newblock}{\relax}
\providecommand{\bibinfo}[2]{#2}
\providecommand{\BIBentrySTDinterwordspacing}{\spaceskip=0pt\relax}
\providecommand{\BIBentryALTinterwordstretchfactor}{4}
\providecommand{\BIBentryALTinterwordspacing}{\spaceskip=\fontdimen2\font plus
\BIBentryALTinterwordstretchfactor\fontdimen3\font minus
  \fontdimen4\font\relax}
\providecommand{\BIBforeignlanguage}[2]{{%
\expandafter\ifx\csname l@#1\endcsname\relax
\typeout{** WARNING: IEEEtran.bst: No hyphenation pattern has been}%
\typeout{** loaded for the language `#1'. Using the pattern for}%
\typeout{** the default language instead.}%
\else
\language=\csname l@#1\endcsname
\fi
#2}}
\providecommand{\BIBdecl}{\relax}
\BIBdecl

\bibitem{pix2pix}
P.~Isola, J.-Y. Zhu, T.~Zhou, and A.~A. Efros, ``Image-to-image translation
  with conditional adversarial networks,'' in \emph{Proceedings of the IEEE
  conference on computer vision and pattern recognition}, 2017, pp. 1125--1134.

\bibitem{cyclegan}
J.-Y. Zhu, T.~Park, P.~Isola, and A.~A. Efros, ``Unpaired image-to-image
  translation using cycle-consistent adversarial networks,'' in
  \emph{Proceedings of the IEEE international conference on computer vision},
  2017, pp. 2223--2232.

\bibitem{UNIT}
M.-Y. Liu, T.~Breuel, and J.~Kautz, ``Unsupervised image-to-image translation
  networks,'' \emph{Advances in neural information processing systems},
  vol.~30, 2017.

\bibitem{Combogan}
A.~Anoosheh, E.~Agustsson, R.~Timofte, and L.~Van~Gool, ``Combogan:
  Unrestrained scalability for image domain translation,'' in \emph{Proceedings
  of the IEEE conference on computer vision and pattern recognition workshops},
  2018, pp. 783--790.

\bibitem{cheung2014discovering}
B.~Cheung, J.~A. Livezey, A.~K. Bansal, and B.~A. Olshausen, ``Discovering
  hidden factors of variation in deep networks,'' \emph{arXiv preprint
  arXiv:1412.6583}, 2014.

\bibitem{kingma2014semi}
D.~P. Kingma, S.~Mohamed, D.~Jimenez~Rezende, and M.~Welling, ``Semi-supervised
  learning with deep generative models,'' \emph{Advances in neural information
  processing systems}, vol.~27, 2014.

\bibitem{mathieu2016disentangling}
M.~F. Mathieu, J.~J. Zhao, J.~Zhao, A.~Ramesh, P.~Sprechmann, and Y.~LeCun,
  ``Disentangling factors of variation in deep representation using adversarial
  training,'' \emph{Advances in neural information processing systems},
  vol.~29, 2016.

\bibitem{infogan}
X.~Chen, Y.~Duan, R.~Houthooft, J.~Schulman, I.~Sutskever, and P.~Abbeel,
  ``Infogan: Interpretable representation learning by information maximizing
  generative adversarial nets,'' \emph{Advances in neural information
  processing systems}, vol.~29, 2016.

\bibitem{DRIT}
H.-Y. Lee, H.-Y. Tseng, J.-B. Huang, M.~K. Singh, and M.-H. Yang, ``Diverse
  image-to-image translation via disentangled representations,'' in
  \emph{European Conference on Computer Vision}, 2018.

\bibitem{MUNIT}
X.~Huang, M.-Y. Liu, S.~Belongie, and J.~Kautz, ``Multimodal unsupervised
  image-to-image translation,'' in \emph{Proceedings of the European conference
  on computer vision (ECCV)}, 2018, pp. 172--189.

\bibitem{seqslam}
M.~J. Milford and G.~F. Wyeth, ``Seqslam: Visual route-based navigation for
  sunny summer days and stormy winter nights,'' in \emph{2012 IEEE
  international conference on robotics and automation}.\hskip 1em plus 0.5em
  minus 0.4em\relax IEEE, 2012, pp. 1643--1649.

\bibitem{siam2017fast}
S.~M. Siam and H.~Zhang, ``Fast-seqslam: A fast appearance based place
  recognition algorithm,'' in \emph{2017 IEEE International Conference on
  Robotics and Automation (ICRA)}.\hskip 1em plus 0.5em minus 0.4em\relax IEEE,
  2017, pp. 5702--5708.

\bibitem{lowe2004distinctive}
D.~G. Lowe, ``Distinctive image features from scale-invariant keypoints,''
  \emph{International journal of computer vision}, vol.~60, no.~2, pp. 91--110,
  2004.

\bibitem{sift}
------, ``Object recognition from local scale-invariant features,'' in
  \emph{Proceedings of the seventh IEEE international conference on computer
  vision}, vol.~2.\hskip 1em plus 0.5em minus 0.4em\relax Ieee, 1999, pp.
  1150--1157.

\bibitem{orb}
E.~Rublee, V.~Rabaud, K.~Konolige, and G.~Bradski, ``Orb: An efficient
  alternative to sift or surf,'' in \emph{2011 International conference on
  computer vision}.\hskip 1em plus 0.5em minus 0.4em\relax Ieee, 2011, pp.
  2564--2571.

\bibitem{bow}
D.~G{\'a}lvez-L{\'o}pez and J.~D. Tardos, ``Bags of binary words for fast place
  recognition in image sequences,'' \emph{IEEE Transactions on Robotics},
  vol.~28, no.~5, pp. 1188--1197, 2012.

\bibitem{vlad}
H.~J{\'e}gou, M.~Douze, C.~Schmid, and P.~P{\'e}rez, ``Aggregating local
  descriptors into a compact image representation,'' in \emph{2010 IEEE
  computer society conference on computer vision and pattern
  recognition}.\hskip 1em plus 0.5em minus 0.4em\relax IEEE, 2010, pp.
  3304--3311.

\bibitem{DenseVLAD}
A.~Torii, R.~Arandjelovic, J.~Sivic, M.~Okutomi, and T.~Pajdla, ``24/7 place
  recognition by view synthesis,'' in \emph{Proceedings of the IEEE conference
  on computer vision and pattern recognition}, 2015, pp. 1808--1817.

\bibitem{netvlad}
R.~Arandjelovic, P.~Gronat, A.~Torii, T.~Pajdla, and J.~Sivic, ``Netvlad: Cnn
  architecture for weakly supervised place recognition,'' in \emph{Proceedings
  of the IEEE conference on computer vision and pattern recognition}, 2016, pp.
  5297--5307.

\bibitem{Benchmarking}
T.~Sattler, W.~Maddern, C.~Toft, A.~Torii, L.~Hammarstrand, E.~Stenborg,
  D.~Safari, M.~Okutomi, M.~Pollefeys, J.~Sivic \emph{et~al.}, ``Benchmarking
  6dof outdoor visual localization in changing conditions,'' in
  \emph{Proceedings of the IEEE conference on computer vision and pattern
  recognition}, 2018, pp. 8601--8610.

\bibitem{porav2018adversarial}
H.~Porav, W.~Maddern, and P.~Newman, ``Adversarial training for adverse
  conditions: Robust metric localisation using appearance transfer,'' in
  \emph{2018 IEEE international conference on robotics and automation
  (ICRA)}.\hskip 1em plus 0.5em minus 0.4em\relax IEEE, 2018, pp. 1011--1018.

\bibitem{todaygan}
A.~Anoosheh, T.~Sattler, R.~Timofte, M.~Pollefeys, and L.~Van~Gool,
  ``Night-to-day image translation for retrieval-based localization,'' in
  \emph{2019 International Conference on Robotics and Automation (ICRA)}.\hskip
  1em plus 0.5em minus 0.4em\relax IEEE, 2019, pp. 5958--5964.

\bibitem{DIFLFCL}
H.~{Hu}, H.~{Wang}, Z.~{Liu}, C.~{Yang}, W.~{Chen}, and L.~{Xie},
  ``Retrieval-based localization based on domain-invariant feature learning
  under changing environments,'' in \emph{2019 IEEE/RSJ International
  Conference on Intelligent Robots and Systems (IROS)}, 2019, pp. 3684--3689.

\bibitem{DISAM}
H.~{Hu}, H.~{Wang}, Z.~{Liu}, and W.~{Chen}, ``Domain-invariant similarity
  activation map metric learning for retrieval-based long-term visual
  localization,'' \emph{IEEE/CAA Journal of Automatica Sinica}, pp. 1--16,
  2021.

\bibitem{tang2020adversarial}
L.~Tang, Y.~Wang, Q.~Luo, X.~Ding, and R.~Xiong, ``Adversarial feature
  disentanglement for place recognition across changing appearance,'' in
  \emph{2020 IEEE International Conference on Robotics and Automation
  (ICRA)}.\hskip 1em plus 0.5em minus 0.4em\relax IEEE, 2020, pp. 1301--1307.

\bibitem{DRIT_plus}
H.-Y. Lee, H.-Y. Tseng, Q.~Mao, J.-B. Huang, Y.-D. Lu, M.~K. Singh, and M.-H.
  Yang, ``Drit++: Diverse image-to-image translation viadisentangled
  representations,'' \emph{International Journal of Computer Vision}, pp.
  1--16, 2020.

\bibitem{gcgan}
H.~Fu, M.~Gong, C.~Wang, K.~Batmanghelich, K.~Zhang, and D.~Tao,
  ``Geometry-consistent generative adversarial networks for one-sided
  unsupervised domain mapping,'' in \emph{Proceedings of the IEEE/CVF
  Conference on Computer Vision and Pattern Recognition}, 2019, pp. 2427--2436.

\bibitem{FAB-MAP}
M.~Cummins and P.~Newman, ``Fab-map: Probabilistic localization and mapping in
  the space of appearance,'' \emph{The International Journal of Robotics
  Research}, vol.~27, no.~6, pp. 647--665, 2008.

\bibitem{cmudata}
H.~Badino, D.~Huber, and T.~Kanade, ``{The CMU Visual Localization Data Set},''
  \url{http://3dvis.ri.cmu.edu/data-sets/localization}, 2011.

\end{thebibliography}
}

\end{document}